# Disparity-based HDR imaging

Jennifer Bonnard    Gilles Valette    Céline Loscos
gilles.valette@univ-reims.fr    celine.loscos@univ-reims.fr
CReSTIC laboratory
University of Reims Champagne-Ardenne

## ABSTRACT

High-dynamic range imaging permits to extend the dynamic range of intensity values to get close to what the human eye is able to perceive. Although there has been a huge progress in the digital camera sensor range capacity, the need of capturing several exposures in order to reconstruct high-dynamic range values persist. In this paper, we present a study on how to acquire high-dynamic range values for multi-stereo images. In many papers, disparity has been used to register pixels of different images and guide the reconstruction. In this paper, we show the limitations of such approaches and propose heuristics as solutions to identified problematic cases.

## 1. INTRODUCTION

Conventional digital images are called LDR (Low Dynamic Range). They correspond to most images acquired so far with cameras SLRs and compact cameras but also with the cameras available in the cell phones for example. They are usually captured and stored in an 8 to 10-bit precision format. When the dynamic color range is extended on at least 16 bits, the images are called HDR (High Dynamic Range) [21]. Digital HDR imaging has existed for more than twenty years and is now well integrated to the general public because its feature is now accessible on mobile phones in particular. HDR imaging is intended to overcome the shortcomings of current sensors while approaching solutions representing the range of intensities perceptible by the human eye. In parallel with digital imaging that can be considered two-dimensional, the three-dimensional images (3D images) enable to perceive the relief of a scene. This technology is based on the geometry defined by stereoscopic vision.

The fields of HDR imaging and 3D imaging have evolved a lot but independently. The development of methods allowing the acquisition or the generation of 3D and HDR images is booming but remains without real solution for the general public. The methods we propose deal with the fusion of the two domains of to obtain 3D HDR images with a wider dynamic range of colors than conventional 3D LDR images.

In a previous approach, we proposed a disparity-based framework for 3D HDR imaging, from acquisition to generation [3]. In this paper, we propose a two-step solution to correct local errors. Our first contribution is an automatic detection of errors. Our second contribution is a set of proposed approaches to generate HDR values for identified problematic pixels.

After reviewing previous work (section 2) and providing more details on our original approach (section 3), we describe our two contributions: automatic error detection (section 4) and HDR value computation for erroneous pixels (section 5). In section 6, we give detailed objective results on tests run on different datasets.

## 2. PREVIOUS WORK

**3D reconstruction by stereovision.** 3D reconstruction by stereovision relates to the automatic depth extraction of a 3D scene structure from different viewpoints (2 to n) acquired at the same time. It comes down to match all homologous pixels from the 3D point projections on the n images. This paper considers simplified multi-epipolar geometry, reached either by using directly a capture configuration in parallel geometry [17] or applying a pre-processing step of rectification on each image [9], leading to epipolar lines parallel to image columns or rows. A matching scheme defines data similarity (or dissimilarity) within a given neighborhood between potential homologous pixels, rendered difficult by lack of information in images (such as occlusions) or ambiguous information (such as homogeneous/repetitive area or luminosity variations). Multiscopic methods usually gain robustness with information redundancy by computing simultaneously the n depth maps [16].

**HDR imaging** Traditional HDR image reconstruction methods [21] combine information captured at different exposure times to acquire different intensities of a scene [6][11]. One difficulty with this approach is to handle dynamic scenes, where objects can move during the acquisition process. Several methods were proposed [11] but very few of them aim at reconstructing HDR values for dynamic environments[19]. Combining HDR with stereovision enables high-quality depth perception reproduction of real-world scenes, but few contributions have been made in this domain. Solutions were proposed for 3D HDR images with stereo cameras [22][10] or multi-stereo cameras [3] following stereovision-based procedures, with remaining inaccuracies in under- or over-exposed areas. This is improved using patch-map along the epipolar line [20] but spatial coherence is lost.

## 3. METHOD OVERVIEW

### 3.1 Original framework for multiscopic HDR reconstruction

In [3], we presented a framework to obtain, from a set of multi-view and multi-exposure LDR images, a set of HDR images, one image being generated for each point of view. The depth information provided by the multi-view aspect of the images is used to find matching pixels in all images. The information provided by the different exposures is then used to obtain a radiance value for each pixel. An example of acquisition and results is given Figure 1. The output of our method can be used as an input for autostereoscopic display after application of a tone-mapping algorithm. Moreover, since a disparity map is produced for each image, augmented reality applications are also conceivable.

Our approach consists of two main steps. First, we identify pixels representing the same point of the scene on different views. For this step, we adapted an approach from Niquin et al. [16] to the multi-exposure context. Second, for each view, we calculate an HDR image, based on the list of matching pixels. We reformulated the well-known method of Debevec and Malik [6] to consider the multi-view aspect of the images.

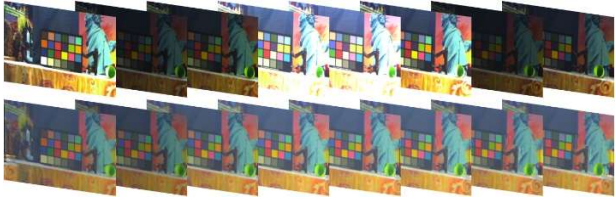

**Figure 1. Example of input (top) and output (bottom) of our method (HDR images tone mapped using [7] ).**

### 3.2  Stereo matching

Usually, the images acquired to obtain an HDR image are captured from a single viewpoint. In this case, the pixels representing the same point in the scene are in the same position in each image. This is no longer the case when the acquisition of the images is carried out from different points of view. These pixels, to which an HDR value is to be assigned, are then at different coordinates in each image where they are represented. A matching step is therefore necessary in order to match the pixels representing the same point in each image constituting the dataset.

We call a *match* a set of matching pixels, i.e., a set of pixels representing the same point of a scene which will get a unique HDR value. Using this notion of match, the method proposed by Niquin et al. [16] aims to simultaneously build a map of disparities for each image and a match list in order to perform a 3D reconstruction of the scene.

It is worth noticing that this stereo matching method is based on color similarity and only processes images with the same exposure. Therefore, we bring the images of different exposures back to a common exposure in order to use the method as is. If the data are linear, a factor based on the percentage of light reaching the sensor is applied to the images. If they are not linear, a first step consists in estimating the response curve associated with the sensor in question. For this, we exploit the method of Debevec and Malik [6].

### 3.3  Multi HDR value computation

In our framework we adapt the commonly used formula of Debevec and Malik [6] for calculating the HDR value (radiance E) associated with a pixel. The initial equation is recalled in Eq. (1). This equation considers the set of points on the scene at the same position (i, j) in the set of images considered. The calculation is carried out separately on each color component.

$$E(i,j) = \frac{\sum_{n=1}^{N} \omega(I_n(i,j)) \left(\frac{f^{-1}(I_n(i,j))}{\Delta t_n}\right)}{\sum_{n=1}^{N} \omega(I_n(i,j))} \quad (1)$$

where is the total number of images, $I_n(i,j)$ the color value of the coordinate pixel (i,j) in the image $I_n$ acquired with an exposure time $\Delta t_n$ et $f^{-1}$ the inverse function of the camera response and ω a weight function for penalizing under- or over-exposed pixels.

We adapted Eq. (1) to allow the computation of a radiance value $E_c$ on the color component c associated with each match m.

$$m = \{ q_i \mid i \in A \subseteq \{1..N\} \},$$

$$E_c(m) = \frac{\sum_{i \in A} \omega(q_{ic}) \left(\frac{f^{-1}(q_{ic})}{\Delta t_i}\right)}{\sum_{i \in A} \omega(q_{ic})} \quad (2)$$

where m is a match, $q_{ic}$ the color value of the pixel $q_i$ belonging to the m match on the color component c ∈ {R,G,B}, $\Delta t_i$ the exposure time of the image $I_i$ to which the pixel $q_i$ belongs, $f^{-1}$ the inverse function of the camera response and ω a weight function.

### 3.4  Proposed extension

We have shown in previous articles [3][4] that the results of our original framework presented objectively quantifiable imperfections. We propose an automatic algorithm to detect invalid pixels (see details in section 4). We identified several possible methods to compute a corrected HDR value on these detected invalid pixels. They are presented in section 5. These additional steps to our original approach are presented in Figure 2.

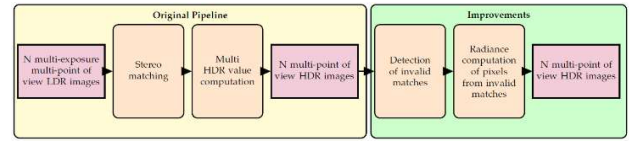

**Figure 2. Extended pipeline: the green box represents the additional steps proposed to improve the radiance of the pixels of invalid matches.**

## 4.  AUTOMATIC DETECTION OF INVALID MATCHES

An invalid match is a match where the radiance of its pixels is considered as incorrect. To automatically detect the invalid matches in a set of HDR images, we based our algorithm on a subjective evaluation that the colors of the pixels belonging to these matches are colored differently than their neighbors, as shown in the examples of the Figure 3.

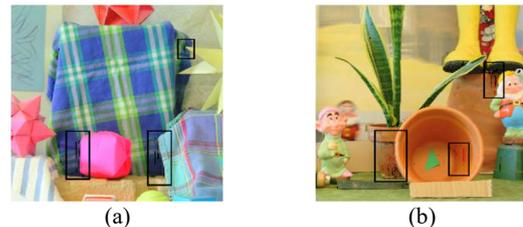

(a)          (b)

**Figure 3. HDR images produced by our method on the first view for the image set (a) Moëbius and (b) Dwarves of the Middlebury database. Black boxes outline pixels inconsistent with their neighborood.**

The determination of the non-acceptance of the radiance of the pixels of a match is based on the sum of the RGB values of each pixel of this match. This sum must belong to an interval [a,b] so that the match is not considered invalid. The limits of this interval depend on the range of LDR values on which the original images are stored, and correspond to a selection of non under- or over-exposed pixels. A second criterion considers the number of different exposures represented in the match under consideration. If the complete set of exposures available in the processed images is not represented, we consider that the radiance of the pixels of this match must be corrected. In this way we can distinguish a dark-colored pixel from an under-exposed pixel and a light-colored pixel from an over-exposed pixel. The set of invalid matches detected is then placed in a list treated by each of the methods of improvement of the radiances proposed in the following sections.

## 5.  HDR VALUE FOR INVALID MATCHES

We propose three solutions to correct the radiance of the invalid pixels detected by the method described in the previous section. The first method permits to assign a new pixel radiance to the current pixel by using the radiance of the pixels in its neighbourhood (see section 5.1). The two other methods are decomposed in two steps: correction of the initial disparities and then correction of the radiances of the pixels. The first method modifies the disparity of the pixels by taking into account the disparity that is attributed to the pixels of the neighborhood while the second method takes into account the disparities calculated on each of the color components R, G and B separately. The new disparity maps lead to new lists of matches: the correction of the radiances of the pixels treated by these two methods is then based on the exploitation of these new list of matches.

## 5.1 Correction by interpolation of radiances

The steps for assigning new radiances to the pixels of invalid matches by the color-based method are detailed in algorithm 1. The method consists in interpolating the HDR values by considering the pixels of the 3×3 neighbourhood of all the pixels of an invalid match. The radiance of the set of pixels of a neighborhood is considered with a confidence index α in order to distinguish between the pixels of a valid match and the pixels of an invalid match. This confidence index can have three values: 0 for invalid matches pixels, 0.5 for invalid matches pixels whose radiance has been corrected and 1 for valid matches pixels. Thus, a pixel of an invalid match can see its confidence index rise from 0 to 0.5 but never to 1, this value being reserved for the pixels of a valid match.

**Algorithm 1:** Assigning new radiances to the pixels of invalid matches.
1: Initialization of a threshold
2: **while** the list of invalid matches is non-empty **and** the sum of the indices of confidence is greater than the threshold **do**
3:   **for** each invalid match **do**
4:     **if** the sum of the confidence indices of the pixels of the neighborhood is greater than the threshold **then**
5:       calculate the new radiance
6:       update the radiance of the pixels the match
7:       set the confidence index α of the pixels the match to 0.5
8:       delete the match from the list of invalid matches
9:     **end if**
10:   **end for**
11:   **if** the size of the list of invalid matches does not change between two iterations of the same threshold **then**
12:     decrease the threshold
13:   **end if**
14: **end while**

The zero confidence index means that as long as the radiance of a pixel from the list of invalid matches has not been corrected, it cannot be considered in the computation of the radiance of another pixel. To treat the pixels whose neighborhood is of better quality, we impose a constraint on the sum of the confidence indices of the neighboring pixels. This constraint is gradually released so that all pixels can be processed. The processing algorithm is iterative in order to manage the priority constraint in the main loop. It implies the impossibility of processing all of the invalid matches in a single iteration. Therefore, the process is repeated until all the radiances of the pixels are replaced. When there is no change for a given confidence index, it is decremented by 0.5. It is only at the end of the current iteration that the radiance values calculated during this iteration are allocated to the pixels concerned.

For a set of 8 multiscopic images, the radiance attributed to the pixels of an invalid match m is obtained by the formula:

$$E(p) = \frac{\sum_{i=1}^{Card(m)} \sum_{j=0}^{7} \alpha_{ij} E_i(p_j)}{\sum_{i=1}^{Card(m)} \sum_{j=0}^{7} \alpha_{ij}} \quad (3)$$

where p is the current pixel of the invalid match processed, $p_j$ is a pixel of the neighborhood of the pixel considered with $j \in \{0, .., 7\}$ its identifier, $E_i(p_j)$ is the radiance of the pixel $p_j$ and $\alpha_{ij}$ corresponds to the confidence index of the pixel $p_j$.

## 5.2 Correction based on disparities

In this section, we propose a solution to correct the disparities originally attributed to the pixels of invalid matches. According to our observations, the disparity of these pixels is wrong because many of them are on the contours of the objects. A bad disparity leads to the pairing of pixels not representing the same point of the scene. Invalid matches cannot be kept as is. The pixels of these matches are separated so that they can be treated separately and are considered as singleton matches.

The proposed methods are divided into three steps (see Figure 4). When the disparity maps are computed, a new pairing of the pixels is carried out. Then we look in this new list of matches which are now the pixels counterparts of the invalid matches. We can then calculate an HDR value only for the pixels of the invalid matches. The radiance of the other pixels remains unchanged.

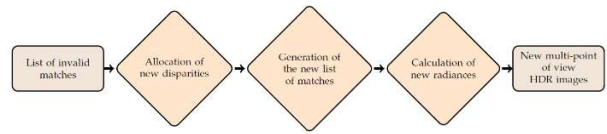

**Figure 4. Pipeline generation of new HDR values from the list of invalid matches.**

Two methods are proposed for correcting disparities. The first one is based on heuristics about the disparities of the neighborhood. The second one performs the computation on monochromatic values.

### 5.2.1 Method based on heuristics on the disparities of the neighborhood

The choice of a method for modifying disparities by heuristics is guided by the knowledge we have of the disparity of each pixel of the images. We assume that, considering a restricted neighborhood, there is little risk of a large difference between the disparities of the pixels. The advantage of methods of estimating disparities is their sensitivity to the texture of the objects of the scene. Their calculation is then more precise than the modification of the colors presented in the previous section. It may then be possible to retrieve the texture details in a scene based on the continuity of the disparities. Horizontal continuities are favored over vertical continuities. If the disparities are equal, the pixels are inside an object, otherwise the pixels are at the edge of an object or on two different objects.

**Algorithm 2:** Assigning new disparities to the pixels of invalid matches.
1: Initialization of a threshold associated with the maximum number of usable pixels required
2: **while** the list of invalid matches is non-empty and the set threshold is greater than 0 **do**
3:   **for** each invalid match **do**
4:     Retrieve in the vicinity of the current pixel the exploitable pixels
5:     **if** the number of pixels considered is greater than the threshold **then**
6:       Selection of the new disparity (see algorithm 3)
7:     **end if**
8:   **end for**
9:   Update pixel disparity of marked matches
10:   Remove marked matches from the list of invalid matches
11:   **if** the size of the invalid match list does not change **then**
12:     Decrease the threshold
13:   **end if**
14: **end while**

The proposed method is summarized in the algorithm 2. Considering the neighborhood of size 3×3, the number of pixels considered decreases from 8 to 1, giving priority to the greatest number of exploitable pixels, then the condition is gradually released. The algorithm is interrupted when the threshold of the number of exploitable pixels becomes zero. This stop criterion implies that the processing loop may stop before all pixels are processed.

The consideration of the neighborhood to determine the new radiance to be allocated to a pixel is dependent on the disparities of the neighboring pixels, which can potentially propagate errors. The recalculated disparities are considered with the same reliability as the initial disparities. The distinction is made between a valid match and an invalid match but a differentiation is also made at the level of the pixels. A pixel can be exploitable or non-exploitable. Its exploitability is based on the LDR value of its color components. If at least one of them is not within a defined range, the pixel is considered to be non-exploitable. Otherwise, it is exploitable and can therefore be used in the proposed methods. In this definition, a pixel of which the set of the under-exposed or over-exposed color components is considered to be non-exploitable.

| $p_0\ d_0$ | $p_1\ d_1$ | $p_2\ d_2$ |
|---|---|---|
| $p_7\ d_7$ | $p\ d$ | $p_3\ d_3$ |
| $p_6\ d_6$ | $p_5\ d_5$ | $p_4\ d_4$ |

**Figure 5. Pixels and disparities labelling of the neighborhood 3×3 of a pixel p of an invalid match.**

The disparity d of the current pixel is calculated as a function of the disparities of the exploitable pixels of the neighborhood. It is defined by the algorithm 3. This choice is guided by the fact that the difficulties of pairing essentially take place at the edges of the objects. Horizontal and then vertical consistencies are favored first in equality and then in increment. If the neighboring disparities do not correspond to these cases, a median value of the disparities is attributed.

```
Algorithm 3: Allocation of new disparities by means of heuristics on neighborhood
disparities (see definition of indices figure 5).
1:  P = {p_i i ∈ [0;7] | p_i exploitable }
2:  if {p_3, p_7} ⊂ P and d_3 = d_7 then
3:      d ← d_3
4:  else if {p_1, p_5} ⊂ P and d_1 = d_5 then
5:      d ← d_1
6:  else if {p_3, p_7} ⊂ P and d_7 < d_3 then
7:      d ← d_7
8:  else if {p_3, p_7} ⊂ P and d_3 < d_7 then
9:      d ← d_3
10: else if {p_1, p_5} ⊂ P and d_1 < d_5 then
11:     d ← d_1
12: else if {p_1, p_5} ⊂ P and d_5 < d_1 then
13:     d ← d_5
14: else
15:     d ← median disparity of exploitable pixels disparities
16: end if
```

The method is based on the assumption that the disparity is wrong, so it is necessary for the new disparity to be different from the initial one. However, this method does not ensure the modification of all the disparities of the pixels of the invalid matches. The new disparity may sometimes be identical to the original disparity without any other choice in the attribution.

*5.2.2  Method based on mono-chromatic disparities*

The second method proposed to modify the disparities of the pixels of invalid matches is based on disparity maps independently computed on each color component R, G, B. Assuming that a pixel may not be under-exposed or over-exposed on at least one of the color components, we exploit this independence so as to modify the disparity of a pixel by taking into account the disparities calculated on the three color components separately. The method consists first of all of considering only one component on each of the multi-exposure images. The same component is chosen for the set of images. Thus, from the RGB images, we obtain a set of images corresponding to the only R component (the same for the G and B components). Each of the three sets of images is therefore used separately to generate the disparity maps associated with each color component.

For each pixel, we now have three disparities obtained respectively from the R images, the G images and the B images. The choice of the disparity is based on the LDR values R, G and B of the processed pixel. A pixel whose color is close to the median value range on which the LDR images are stored is neither underexposed nor overexposed. Its disparity is therefore likely to have been correctly calculated. We classify the values R, G and B as a function of their distance from the median. The disparity chosen is that calculated from the color component whose distance to the median is the lowest. We verify that this disparity is different from the initial disparity. If this is not the case, the choice is the disparity calculated on the second component whose value is closest to the median, and so on. As with the previous method, it is possible that the new disparity cannot be different from the initial disparity. In this case, the correction is not made and the initial disparity is retained.

## 6.  RESULTS AND DISCUSSIONS
### 6.1  Image dataset

We tested our approaches on three different datasets. The first one is generated using the OctoCam multiview camera [17] equipped with eight horizontally aligned and synchronized objectives designed to deliver 3D content for auto-stereoscopic displays. This camera is based on a simplified epipolar geometry that permits strong assumptions on 3D stereovision algorithms [18] and horizontally aligned epipolar lines. Each of its sensors allows the acquisition of 10 bits per color channel. A neutral density filter is fixed on each objective; consequently, a different percentage of the light reaches the sensor for each view, hence acquired images represent different exposures.

The second series of images was generated using the POV-Ray ray-tracer. We reproduced the geometry of the OctoCam to render eight images from aligned viewpoints of a synthetic scene.

The third source of images is the database made available on the Internet by Middlebury University[1] which offers images acquired from different points of view and different exposures. Contrary to those of OctoCam, the images proposed by Middlebury are acquired according to a parallel geometry so it was necessary to choose a region of interest in the original images in order to satisfy the requirements of the off-centered parallel geometry.

### 6.2  Objective quality metrics

To judge on the quality of an HDR image, the most common method is to compare it with a reference image. The HDR-VDP (High Dynamic Range Visible Difference Predictor), developed by Mantiuk et al.[13] has been developed for this type of image. An update of this metric was proposed by Narwaria et al. [14]. One of the major disadvantages of this predictor is the number of parameters to be considered. However, for simplicity a version is available online[2]. On the same principle, the authors have developed a method for HDR videos[15]. Aydin et al. [1] and Valenzise et al. [23] proposed comparing two HDR images with the measures traditionally used to compare two LDR images such as PSNR (Peak Signal-to-Noise Ratio) and SSIM (Structural Similarity Index) by adapting the data used. Instead of taking into account the color components Red, Green and Blue, the data is transformed to become perceptually uniform. When the data were not linear, Narwaria et al. [14] demonstrated better HDR-VDP performance.

---

[1] http://vision.middlebury.edu/stereo/data/

Hanhart et al. [8] conclude that HDR-VDP-2 and PQ2VIFP are the best generic predictors of visual quality as they show less content dependency than the other metrics.

### 6.3 Evidence of limitation on objects' edges

Visual errors occur because of the difficulty of aggregating enough pixels in a match to obtain a correct HDR value. Whatever the set of images considered, errors are mostly localized at the contours of the objects since there are lower numbers of correspondents. Figure 6 presents the image acquired on the view #0 for the Dolls image set of the Middlebury database and the number of pixels contained in the match to which the current pixel belongs. In the neighborhood of objects' edges, the number of pixels contained in the matches decreases.

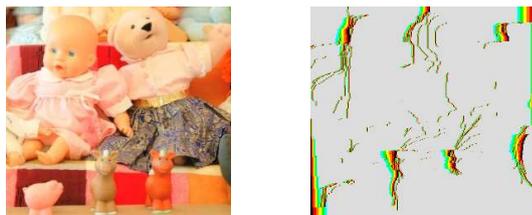

**Figure 6. Distribution of the number of pixels per match in the HDR image generated on the view #0 of the Dolls image set. The associated color code is: gray-7 pixels, cyan-6 pixels, green-5 pixels, yellow-4 pixels, orange-3 pixels, red-2 pixels and black-1 pixel.**

### 6.4 Evaluation of the method of automatic detection of invalid matches

A comparison is made between the reference HDR images and the HDR images produced by our uncorrected method. This comparison is made independently on each of the points of view. A threshold, fixed at 6% of the range of pixel radiance values, is considered to be tolerable. The pixels whose Euclidean distance in the RGB space is greater than this threshold are considered incorrect. We obtain reference values on the number of correct pixels and the number of incorrect pixels. These values are compared to those obtained by our automatic detection.

Table 1 shows a series of percentages calculated on the images acquired with the OctoCam when the neutral density filters are fixed on some of the camera lenses and on the image set Dwarves of the Middlebury database. The Correct Pixels column binds the number of pixels correctly classified as correct by our method to the correct number of pixels in the reference method. The Incorrect Pixels column corresponds to the number of pixels correctly classified as incorrect by the proposed detection method relative to the number of incorrect pixels in the reference method. The False Positives column represents the number of pixels that are incorrectly classified as incorrect by our method relative to the total number of pixels that are classified as incorrect by the same method. The False Negatives column relates the number of pixels that are incorrectly classified as correct by the proposed method and the total number of pixels that are classified as correct in the same method.

**Table 1. Distribution of the detection of invalid matches in comparison with the difference between the reference HDR image and the generated HDR image.**

| Vue | Correct Pixels | | Incorrect Pixels | | False Positives | | False Negatives | |
|---|---|---|---|---|---|---|---|---|
| | Filtre | Dwarves | Filtre | Dwarves | Filtre | Dwarves | Filtre | Dwarves |
| 0 | 99.20 | 99.25 | $1.32 \cdot 10^{-2}$ | 7.82 | 100 | 35.41 | 4.65 | 14.04 |
| 1 | 99.65 | 99.37 | $3.09 \cdot 10^{-2}$ | $4.05 \cdot 10^{-2}$ | 100 | 99.36 | 4.69 | 9.09 |
| 2 | 99.99 | 99.99 | $4.4 \cdot 10^{-3}$ | 0.27 | 100 | 33.73 | 5.03 | 6.79 |
| 3 | 100 | 99.97 | 0.00 | $1.6 \cdot 10^{-2}$ | 0.00 | 96.82 | 4.75 | 4.99 |
| 4 | 100 | 99.98 | 0.00 | 0.51 | 0.00 | 34.50 | 5.67 | 5.51 |
| 5 | 99.87 | 98.81 | 0.00 | $6.02 \cdot 10^{-2}$ | 100 | 99.58 | 6.89 | 7.84 |
| 6 | 99.47 | 98.59 | 0.00 | 11.51 | 100 | 39.35 | 5.81 | 14.52 |
| 7 | 98.71 | | $8.8 \cdot 10^{-3}$ | | 99.95 | | 4.83 | |

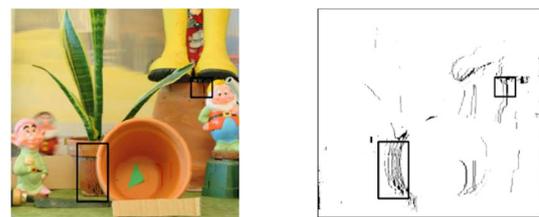

**Figure 7. Erroneous pixels automatically detected by our presented approach.**

These results show that the incorrect pixels are less numerous by our method than by the reference method but they correspond to pixels whose radiance must absolutely be corrected. The reference method needs a threshold whose value is difficult to choose. Our experimentations show that the increase of this threshold increases the number of false positives and reduces the number of false negatives in the proposed method, and its reduction does not allow the detection of clearly identified pixels that are erroneous in the HDR images produced. An example case is shown in Figure 7.

### 6.5 Evaluation of the quality of the generated HDR images

For our work we have chosen to use HDR-VDP-2 (see Figure 8). However, in order to apply this metric, we need HDR reference images to compare to. To obtain the best possible reference images, we generate independently per-viewpoint reference HDR images by combining several exposures of a same viewpoint using the weighted average method of Debevec and Malik [6]. We use three exposures for Middlebury database sets and four for sets acquired with OctoCam and synthetic sets.

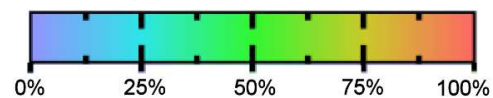

**Figure 8. Color scale of visibility difference probability detection used by HDR-VDP-2.**

Among the three correction methods proposed, the color-based method is the one that produces the higher quality HDR images for the images of the Middlebury database. Figure 9 shows particular

---

[2] http://driiqm.mpi-inf.mpg.de

areas which clearly highlight the improvements made locally by the color-based method.

As shown inFigure 10 and Figure 11, many of the perceptible defects in the HDR images have been corrected. The images produced by the HDR-VDP-2 show a larger blue color range with this method on views 0 and 6, which proves that human eye does not perceive any difference with the reference image in this area. By considering the neighborhood of a pixel, it is possible to assign an HDR value to each pixel, which is not allowed by the original method, because of the consideration of a saturation threshold on the pixels of a match, neither by the two methods based on disparities, since for some pixels the disparity cannot be modified.

For the data set acquired with OctoCam (Figure 12) and the synthesis images (Figure 13), we can see that the results of the methods proposed are close. Using the HDR-VDP-2, we can see some differences between the images produced by each of the methods and those produced by the method without correction, but the pixels concerned are few in number. These two sets of images have fewer automatically detected pixels, which results in the correction of a small number of pixels in the whole image. Therefore, the differences between the methods are not perceived.

Methods based on the improvement of the disparities do not make it possible to achieve the quality of the HDR images produced by the color-based method. The choice to change the disparities is based on the desire to reduce the number of invalid matches to the maximum but these two methods encounter two difficulties. The first is the need to find a disparity different from that originally attributed, which is not always possible since the algorithms may lead to the choice of the same disparity. The second difficulty is, despite the correction of the disparities, to be able to assign a new correct HDR value.

The changes made by the color-based method are more important on the extreme views because the pixels belonging to invalid matches are on these images. Indeed, it is on these views that are placed the images of lower exposures, which makes their pixels more vulnerable. Few pixels are erroneous on the central views so the images appear to be identical to their initial value.

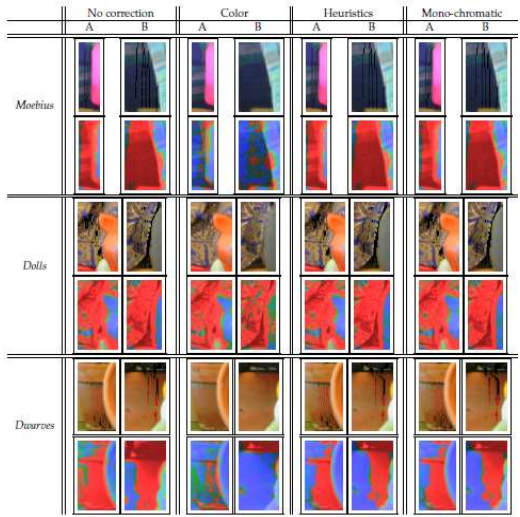

**Figure 9.** HDR and HDR-VDP-2 images corresponding to selected areas in the Middlebury database images for color based, heuristic disparity and mono-chromatic methods.

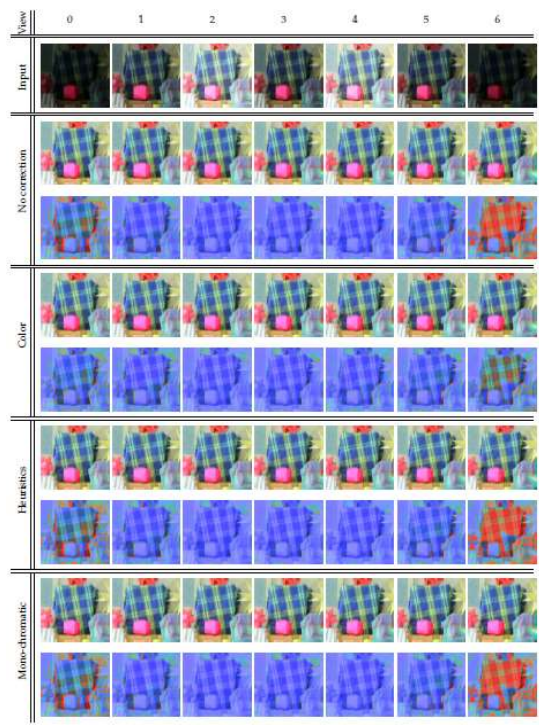

**Figure 10.** HDR and HDR-VDP-2 images on each point of view for color-based, heuristic-disparity and mono-chromatic methods for the Moebius image set of the Middlebury database.

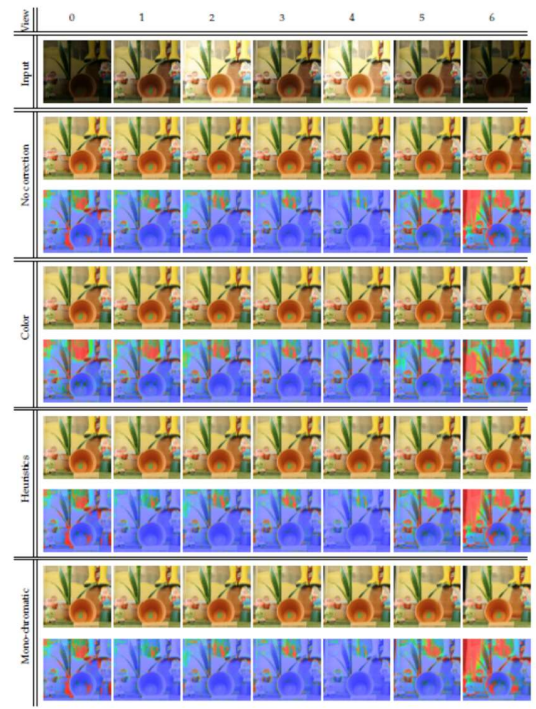

**Figure 11.** HDR and HDR-VDP-2 images on each point of view for the color-based, heuristic-on-disparity and mono-chromatic methods for the Dwarves image set of the Middlebury database.

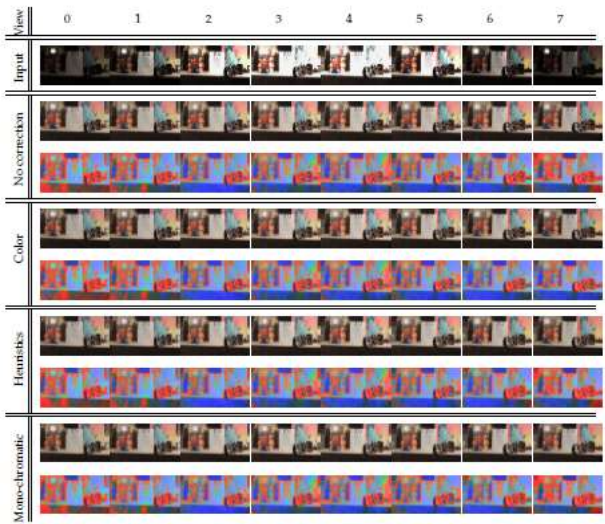

**Figure 12. HDR and HDR-VDP-2 images on each point of view for color-based, heuristic-on-disparity and mono-chromatic methods for the image set acquired with OctoCam mounted with neutral density filters.**

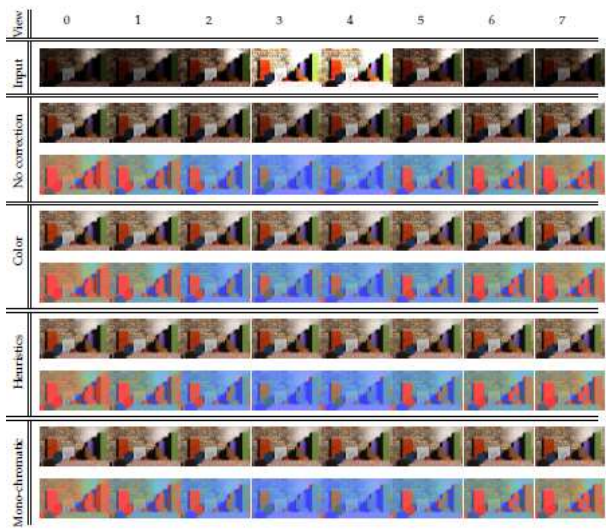

**Figure 13. HDR and HDR-VDP-2 images on each point of view for color-based, heuristic-on-disparity and mono-chromatic methods for the image set generated with the POV-Ray raytracing software.**

### 6.6 Highlighting errors due to the pixel mapping method

The studies we carried out on the geometric and colorimetric coherence of the data acquired for the generation of 3D HDR images [4] do not completely justify the presence of all the errors.

In order to highlight the role of the wrong results in the matching method, new reference images are computed. For these images, the disparity maps and the associated match lists are calculated from the same exposure images, which gives optimal disparity maps. An HDR image per point of view is then calculated from these new elements. This image becomes a new reference that we compare to the reference image used in previous comparisons (see Figure 14). The central image really emphasizes that the errors in the methods come from the method of disparities and not from its application on images of multiple exposures.

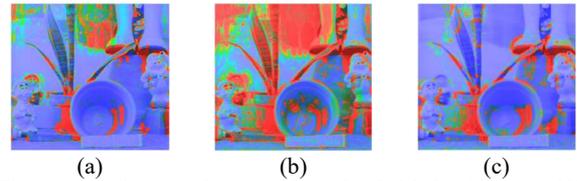

(a) (b) (c)

**Figure 14. Comparisons: a) of the initial reference HDR image with the new reference HDR image, b) of the HDR image with the initial reference HDR image, c) of the HDR image with the new reference HDR image.**

The results of the color-based method clearly show that corrections have been made in areas where errors are due to multi-exposure frames B. The HDR-VDP-2 shows that the corrections made are generally better since the zones A are also improved. It is interesting to note, however, in Box C, errors reintroduced by the method appear (see Figure 15). The radiance attributed to them becomes of lower quality than the initial radiance.

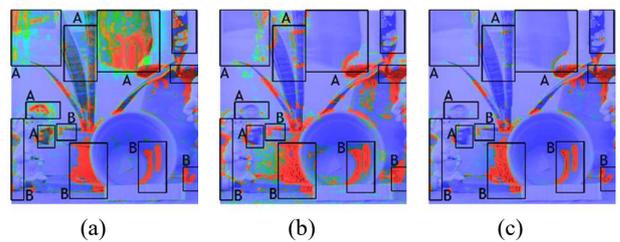

(a) (b) (c)

**Figure 15. Comparisons: a) of the initial reference HDR image with the new reference HDR image, b) of the HDR image with the new reference HDR image, c) of the HDR image after color-based correction with the new reference HDR image.**

### 7. CONCLUSION

In this paper, we address the difficult topic of HDR generation for multiscopic data. The main difficulties, highlighted in this research work, come from the fact that HDR generation requires accurate registration of pixel. However, multiscopic images present non-linear displacement, making thus this registration process difficult. Disparity is explored as a registration approach. We provide an indepth analysis on the limitation of disparity-based HDR generation. One difficulty comes from the fact that disparity solvers rely on color matching whereas data input in our case are differently exposed. Another difficulty is intrinsic to disparity algorithm, where object borders or outliers (highlights) have always been known as difficult to address.

In this paper, we propose an automatic method to locally detect wrongly generated HDR values and we propose correction approaches. Our results show that we are able to correctly identify erroneous pixels and that we are able to significantly improve the results.

As future work, we would like to explore solutions that solve together disparity and HDR values, thus beneficiating both on depth and HDR knowledge. Akhavan et al. [1] demonstrated the benefice of HDR imaging to disparity computation.

### 8. ACKNOWLEDGMENT

This work was funded by the ANR project ReVeRY ANR-17-CE23-0020.